\pdfoutput=1

\documentclass[11pt]{article}

\usepackage{acl}

\usepackage{times}
\usepackage{latexsym}

\usepackage[T1]{fontenc}

\usepackage[utf8]{inputenc}

\usepackage{microtype}

\usepackage{inconsolata}

\usepackage{booktabs}
\usepackage{amssymb}
\usepackage{amsmath}
\usepackage{multirow}
\usepackage{graphicx}
\usepackage{float}
\usepackage{makecell}
\usepackage{longtable}
\usepackage{array}
\usepackage{lscape}
\usepackage{geometry}
\usepackage{tabularx}
\usepackage{xcolor}
\usepackage{colortbl}

\usepackage{hyperref}
\hypersetup{hidelinks,
	colorlinks=true,
	pdfstartview=Fit,
	breaklinks=true}

\newcommand{\moclp}{\textsc{MoCL-P}}

\title{Learn it or Leave it: Module Composition and Pruning \\ for Continual Learning}

\author{
Mingyang Wang$^{1,2,3}$ \hspace*{0.2cm}
{Heike Adel$^{4}$} \hspace*{0.2cm} 
{Lukas Lange$^{1}$} \\
 {\bf Jannik Str\"{o}tgen$^{5}$ \hspace*{0.2cm} Hinrich Sch\"{u}tze$^{2,3}$} \\
  $^1$Bosch Center for Artificial Intelligence, Renningen, Germany \\
  $^2$LMU Munich, Germany \hspace*{0.2cm}
  $^3$Munich Center for Machine Learning (MCML) \\
  $^4$Hochschule der Medien, Stuttgart, Germany \\
  $^5$Karlsruhe University of Applied Sciences, Germany \\
  \texttt{mingyang.wang2@de.bosch.com} 
  }

\begin{document}

\maketitle

\begin{abstract}
In real-world environments, continual learning is essential for machine learning models, as they need to acquire new knowledge incrementally without forgetting what they have already learned.
While pretrained language models have shown impressive capabilities on various static tasks, applying them to continual learning poses significant challenges, including avoiding catastrophic forgetting, facilitating knowledge transfer, and maintaining parameter efficiency. In this paper, we introduce \moclp, a novel lightweight continual learning method that addresses these challenges simultaneously. Unlike traditional approaches that continuously expand parameters for newly arriving tasks, \moclp\ integrates task representation-guided module composition with adaptive pruning, effectively balancing knowledge integration and computational overhead. Our evaluation across three continual learning benchmarks with up to 176 tasks shows that \moclp\ achieves state-of-the-art performance and improves parameter efficiency by up to three times, demonstrating its potential for practical applications where resource requirements are constrained.
\end{abstract}

\section{Introduction}
\label{sec:introduction}

Continual learning (CL) is a learning paradigm aiming at incrementally acquiring and integrating new knowledge over time without forgetting existing knowledge. This capability is essential for machine learning models to stay effective as they encounter dynamic and evolving real-world environments. 
While pretrained language models (PLMs) have demonstrated remarkable capabilities on various static tasks, adapting them for continual task learning remains challenging.

In particular, there are three notable challenges for continual learning. (1) Avoiding catastrophic forgetting: The newly learned information should not disrupt and degrade previously acquired knowledge \cite{mccloskey1989catastrophic}. (2) Facilitating knowledge transfer: The knowledge from past tasks should be reused for efficient learning of new tasks. (3) Maintaining parameter efficiency: The language models need to stay lightweight and effective even if the continual learning sequence scales to hundreds of tasks. 


To mitigate catastrophic forgetting, a line of prior works adopt the idea of \textit{parameter isolation} \cite{razdaibiedina2022progressive, wang2023orthogonal, wang-etal-2023-rehearsal, wang-etal-2024-rehearsal}, 
which allocates isolated parameters dedicated for each task to avoid inter-task interference.
While parameter isolation typically does not allow knowledge transfer across tasks \cite{wang2023orthogonal, wang-etal-2023-rehearsal}, there are attempts to address both challenges of catastrophic forgetting and knowledge transfer at the same time, e.g., by progressively concatenating \cite{razdaibiedina2022progressive} or composing task-specific modules \cite{wang-etal-2024-rehearsal}.


Despite their effectiveness in terms of task performance, parameter isolation methods do not scale well with the number of tasks. When the 
number of tasks in a continual learning sequence is growing into the hundreds, the progressive expansion of task-specific parameters leads to parameter inefficiency and significantly increases computational and storage costs.

In this paper, we address all three continual learning challenges simultaneously and introduce \moclp, a lightweight continual learning approach that leverages task representation-guided module composition and adaptive pruning. First, to avoid catastrophic forgetting, \moclp\ continually adds task-specific modules to PLMs for learning new tasks while keeping the modules frozen once the training on the respective tasks is finished. In addition, to enable knowledge transfer across tasks, \moclp\ allows the model to reuse existing knowledge via module composition. Finally, to keep the language model lightweight, \moclp\ adopts an adaptive pruning strategy by removing modules with redundant information and retaining only the most salient modules throughout the continual learning process.

In our evaluation on three popular datasets as continual learning benchmarks with up to 176 tasks in the learning sequence, \moclp\ stands out by not only showing state-of-the-art performance but also outperforming prior algorithms in parameter efficiency by up to three times across benchmarks.

To the best of our knowledge, this is the first paper that tackles the three  challenges of continual learning simultaneously: \moclp\ avoids catastrophic forgetting, allows knowledge transfer and ensures parameter efficiency. Thus, \moclp\ proposes a sustainable way for continual learning, allowing models to remain lightweight and effective as they evolve with accumulating tasks. 


The code base for \moclp\ is available online.\footnote{https://github.com/boschresearch/MoCL-Pruning}

\section{Related Work}
\label{sec:related-work}
\subsection{Avoiding Catastrophic Forgetting in Continual Learning}
A major challenge in continual learning is known as catastrophic forgetting, where newly learned information disrupts and degrades previously acquired knowledge \cite{mccloskey1989catastrophic}. Existing 
approaches to overcome this issue
can be broadly divided into three categories \cite{wang2023comprehensive}: (1) \textit{Regularization}-based methods explicitly add regularization terms to the loss function to restrict model updates and preserve existing knowledge \cite{li2017learning, kirkpatrick2017overcoming, aljundi2018memory}; (2) \textit{Rehearsal}-based methods leverage a memory buffer to store real examples \cite{rebuffi2017icarl, rolnick2019experience, zhang-etal-2022-continual} or generated pseudo-examples of past tasks for future rehearsal to avoid catastrophic forgetting \cite{shin2017continual, su2019generative}; (3) \textit{Parameter isolation}-based methods construct task-specific parameters to prevent inter-task interference by either dynamically expanding model capacity or isolating existing model weights \cite{madotto2020continual, zhang2022continual, razdaibiedina2022progressive, wang-etal-2023-rehearsal, wang2023orthogonal, wang-etal-2024-rehearsal}.  

Our method, \moclp, belongs to the parameter-isolation based category. We use task representation-guided module composition and adaptive pruning to effectively manage isolated parameters.

\begin{figure*}[h]
    \centering
    \includegraphics[width=1\linewidth]{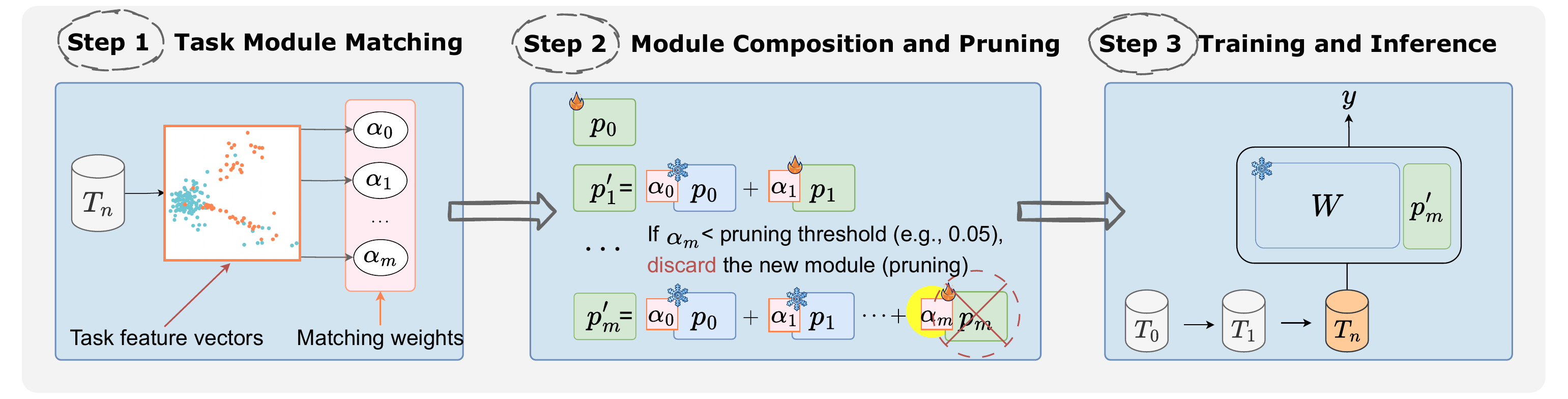}
    \caption{Overview of our proposed method \moclp\ for parameter-efficient continual learning. \textbf{Step 1}: We match the $n$-th task input with task feature vectors to determine the contribution of each existing module for learning the current task. \textbf{Step 2}: We compose the newly initialized module with existing ones and perform adaptive module pruning to preserve only the dominant modules. \textbf{Step 3}: Finally, we combine the composed module $p’_m$ with the PLM for training and inference.}
    \label{fig:main-figure}
\end{figure*}

\subsection{Transferring Knowledge in Continual Learning}
Recent studies in continual learning demonstrate the effectiveness of parameter isolation methods in avoiding catastrophic forgetting \cite{razdaibiedina2022progressive, wang-etal-2023-rehearsal, wang2023orthogonal, wang-etal-2024-rehearsal}. However, naive parameter isolation methods do not allow knowledge transfer across tasks, which leads to inefficient learning as the model cannot leverage previously acquired knowledge to facilitate learning new tasks. To address this, \newcite{yoon2017lifelong} and \newcite{zhu-etal-2022-continual} attempt to first identify reusable modules and only add new parameters when necessary. \newcite{ke-etal-2021-adapting} and \newcite{wang2022dualprompt} introduce knowledge-sharing modules to facilitate knowledge transfer while maintaining task-specific parameters to prevent interference. \newcite{razdaibiedina2022progressive} progressively concatenate task-specific modules to incrementally build a composite model that leverages both new and existing knowledge. \newcite{wang-etal-2024-rehearsal} introduce a modular and compositional continual learning framework to compose the new module with existing ones based on task module matching.

\subsection{Parameter-Efficient Continual Learning}
With the ever-increasing number of parameters in PLMs, it becomes increasingly important to develop machine learning systems that are more scalable, practical, and resource-efficient. In the context of continual learning, this necessitates parameter-efficient approaches that can effectively integrate new knowledge without excessive computational and storage costs as the number of tasks in the continual learning sequence increases.

Recent advancements in continual learning integrate parameter isolation with parameter-efficient fine-tuning (PEFT), i.e., they allocate task-specific PEFT modules for learning and inference \cite{razdaibiedina2022progressive, wang-etal-2023-rehearsal, wang2023orthogonal, wang-etal-2024-rehearsal}. Various PEFT techniques, such as adapter tuning \cite{houlsby2019parameter}, prefix tuning \cite{li2021prefix}, and LoRA \cite{huang-2022-easy}, have been applied in continual learning. Although they reduce the number of training parameters to some extent by freezing the PLM and only updating the PEFT module parameters, it remains challenging to apply them to long-sequence benchmarks that consist of hundreds of tasks. The continuous expansion of task-specific modules leads to significant computational overhead as the number of tasks increases.

Our approach builds on the idea of \newcite{wang-etal-2024-rehearsal} by utilizing task representations for module composition, ensuring that the model effectively reuses relevant knowledge from previous tasks. Beyond that, we introduce an adaptive pruning strategy to keep the language model lightweight and effective throughout the continual learning process, thus making it scalable for continual learning scenarios with long task sequences. 

\section{Problem Definition}
\label{sec:cl-basics}
Continual learning focuses on addressing a series of tasks which arrive in a sequential order. The primary goal is to optimize the model’s average performance across all tasks after learning them sequentially. Formally, the sequence of tasks is denoted as $\{T_1, \dots, T_N\}$. Each task contains a set of input samples $ \{(x^{i}_{n}, y^{i}_{n})\}$. For the text classification tasks we study in this work, $x^{i}_{n}$ is the input text, $y^{i}_{n}$ is the ground-truth label, and $n  \in \{1, \dots, N\}$ is the task identity.

In this work, we focus on rehearsal-free continual learning, i.e., data from earlier tasks is not available when training later tasks. Therefore, our model does not suffer from the memory or privacy issues associated with rehearsal-based methods. We assume the task labels are provided during both training and testing, i.e., task-incremental continual learning \cite{wang2023comprehensive}. However, \moclp\  can be adapted for class-incremental learning, where the task labels are not given during testing, with minor modifications following \newcite{wang-etal-2024-rehearsal}. We leave the exploration of other continual learning settings for future work.

\section{Method}
\label{sec:method}

In this section, we describe \moclp, our proposed CL approach for language models, as illustrated in Figure~\ref{fig:main-figure}, which tackles catastrophic forgetting and enhances knowledge transfer with superior parameter efficiency at the same time.

\subsection{Continual Learning with PEFT}
\label{sec:method-cl-peft}
We inherit the idea of parameter isolation with parameter-efficient fine-tuning (PEFT) introduced in prior work \cite{razdaibiedina2022progressive,wang2023orthogonal, wang-etal-2023-rehearsal, wang-etal-2024-rehearsal}, which allocates trainable PEFT parameters for each task while keeping other parameters frozen.
 
We utilize \textit{prefix-tuning} \cite{li2021prefix} as the PEFT module in consistency with prior works.\footnote{Other PEFT methods like Adapter \cite{houlsby2019parameter} and LoRA \cite{hu2021lora} can also be combined with \moclp\ in general. We leave such exploration for future work.} For each task in the CL sequence, we add a set of trainable PEFT parameters, i.e., a task-specific module, to the pretrained language model (PLM) for downstream task fine-tuning. Instead of updating the whole model, only a small number of the PEFT parameters are optimized. Once training on one given task is completed, the corresponding PEFT module is frozen to preserve the task-specific knowledge in the subsequent training process, thus avoiding catastrophic forgetting.

\subsection{Task Representation-Guided Module Matching}
\label{sec:method-representation-matching}
In contrast to completely isolating task-specific parameters during continual learning, which excludes knowledge transfer, we follow the idea of task module composition introduced in \newcite{wang-etal-2024-rehearsal} to facilitate knowledge transfer.

To this end, we utilize task representations for task module matching, and consequently for composing old and new modules for learning. The module matching aims to determine the contribution of each existing module to learning the current task, i.e., to what extent previously learned modules can be reused for the current task.
 
We introduce trainable feature vectors $V \in \mathbb R^{N \times D}$ as task representations to capture the features of each task in the CL sequence.\footnote{Note that \moclp\ is agnostic to different types of task representations. In addition to the trainable feature vectors, other static task representations such as task embeddings or Gaussian task distributions can also be combined with \moclp. We analyze these options in Section~\ref{sec:res-rep-comparison}.} We set the dimension of each task feature vector $v \in \mathbb R^D $ to the same value as the dimension of the input embeddings $x_n \in \mathbb R^D$. Then, we calculate the cosine similarity between the input embeddings  $x_n$ and each feature vector $v_i$ up to the current task as the matching score $\alpha_i = \cos (x_n, v_i)$. Consequently, we get the module matching weights $\{\alpha_0, \alpha_1, ...\}$ for module composition (details will be introduced in Section~\ref{sec:method-composition-pruning}) to reuse existing knowledge.

\subsection{Module Composition with Adaptive Pruning}
\label{sec:method-composition-pruning}

When the CL learning sequence scales to dozens or hundreds of tasks, the need for efficiency increases. Continuously expanding the module pool to assign a PEFT module to each task, as done in prior works \cite{wang-etal-2023-rehearsal, razdaibiedina2022progressive, wang-etal-2024-rehearsal}, leads to large computational costs. In contrast, 
we employ an adaptive pruning strategy to make our approach scalable in  scenarios with long task sequences.

In particular, our pruning strategy aims at preserving only those modules that add new and valuable information to the set of already selected modules. 
Given a set of selected modules $\{P_0, \dots, P_{m-1}\}$ from previous tasks and a new task $T_n$, $(m-1 \ll n)$, we initialize a trainable module $P_m$ and add it temporarily to the model. For each instance\footnote{For simplicity, we refer to this as $x_n$ in the following.} $x_n^i$ of the current task $T_n$, we compute the matching weights $\{\alpha_0, \dots, \alpha_{m}\}$ by matching $x_n$ with all task feature vectors $\{v_0, \dots, v_m\}$ from our current set of modules. Specifically, we calculate the cosine similarity between $x_n$ and $\{v_0, \dots, v_m\}$ as module matching weights $\alpha_{0:m}$ as detailed in Section \ref{sec:method-representation-matching}. 
Then, we compose the new and old modules via a weighted sum: $P'_m = \sum_{k=0}^{m} \alpha_k P_{k}$. Finally, the composed module $P'_m$ is combined with the PLM, consisting of all the selected module components up to the current task.


After the training on $T_n$ is finished (specifically, the training of the PEFT module $P_m$ and the task feature vector $v_m$), we compare $\alpha_m$, the matching weight of the new module $P_m$, with a threshold\footnote{The threshold is a tunable hyperparameter.} to decide whether to prune $P_m$ or leave it in the set of existing modules. The intuition is that large matching weights indicate new and valuable information, while task modules with small matching weights do not contribute new information and, thus can, be discarded.


\subsection{Training and Inference}
\label{sec:method-train-infer}
The training objective for the $n$-th task in the continual learning sequence is to find the PEFT module $P_{m}$ and the task feature vector $v_m$ that  minimize the cross-entropy loss of training examples, and, at the same time, maximize the cosine similarity between the task-specific feature vector $v_{m}$  and the corresponding task input embeddings $x_n$:

\begin{small}
\begin{equation*}
\min_{P_{m}, v_{m}} - \sum_{x_n, y_n}\log p(y_n | x_n, P'_{n}, \theta) - \sum_{x_n}\cos (x_n, v_{m})
\end{equation*}
\end{small}

Here $P'_n = \sum_{k=1}^{m} \alpha_k P_{k}$ is the weighted summation of the new trainable task module and the existing frozen task modules as introduced in Section~\ref{sec:method-composition-pruning}.
During inference, \moclp\ performs per-instance task module matching and composition. The resulting module is combined with the PLM for inference.

\begin{table*}[!t]
  \centering
  \scalebox{0.9}{
    \begin{tabular}{l|ccccc|ccccc}
    \toprule
    \multicolumn{1}{c|}{\multirow{2}[2]{*}{\textbf{Method}}} & \multicolumn{5}{c|}{\textbf{AfriSenti}} & \multicolumn{5}{c}{\textbf{WikiAnn}} \\
          & \textbf{AVG} & O1 & O2 & O3 & \multicolumn{1}{l|}{\textbf{\# Params}} & \textbf{AVG} & O1 & O2 & O3 & \multicolumn{1}{l}{\textbf{\# Params}} \\
    \midrule
    Seq FT (F) & 6.17  & 5.62  & 6.52  & 6.30  & 560M  & 14.50 & 3.44  & 23.36 & 16.70 & 110M \\
    Seq FT & 49.10 & 50.05 & 49.74 & 47.53 & 0.4M  & 67.99 & 68.25 & 65.05 & 70.66 & 0.1M \\
    Per-task FT & 52.41 & 52.41 & 52.41 & 52.41 & 4.5M  & 71.22 & 71.22 & 71.22 & 71.22 & 24.8M \\
    ProgPrompt & 49.07 & 50.16 & 46.74 & 50.30 & 4.5M  & 73.20 & 73.24 & 73.22 & 73.15 & 24.8M \\
    EPI   & 43.10 & 41.49 & 42.65 & 45.16 & 4.5M  & 67.34 & 67.72 & 67.12 & 67.18 & 24.8M \\
    MoCL  & 59.31 & 59.56 & 58.98 & 59.40 & 4.5M  & 73.80 & 73.78 & 73.81 & 73.82 & 24.9M \\
    \textbf{\moclp\ (Ours)} & \textbf{59.41} & 59.52 & 58.97 & 59.76 & $\text{2.2M}\scriptstyle \pm\text{0.4}$ & \textbf{73.91} & 73.94 & 73.94 & 73.86 & $\text{8.0M}\scriptstyle \pm\text{0.1}$ \\
    \bottomrule
    \end{tabular}%
    }
  \caption{Summary of the continual learning results on two multilingual benchmarks: AfriSenti and WikiAnn NER, with 12 and 176 languages in the task sequence, respectively. We compare \moclp\ with various baseline methods, and show the average model performance (AVG) across different task orders (O1, O2, O3) with the number of trainable parameters (\# Params) used by each method. \moclp\ outperforms other methods with significantly fewer parameters, demonstrating its superiority in both model performance and parameter efficiency in long-sequence continual learning scenarios.}
  \label{tab:overall-1}%
\end{table*}%

\section{Experimental Setup}
In this section, we describe datasets, training details and baselines for our experiments.

\subsection{Datasets}
To evaluate the performance of our method and the effectiveness of its module pruning functionality, we experiment with three continual learning benchmarks, each with long task sequences. Following prior work \cite{razdaibiedina2022progressive, wang-etal-2024-rehearsal}, we use MTL15, a multi-task continual learning benchmark comprising 15 classification tasks, and AfriSenti \cite{muhammad2023afrisenti}, a multilingual sentiment analysis dataset that includes 12 low-resource African languages. Additionally, we include WikiAnn \cite{pan-etal-2017-cross}, a multilingual named entity recognition (NER) dataset covering 176 languages; its long task sequence provides an adequate testbed for the pruning ability of our approach.

We report macro-weighted F1 scores on the AfriSenti benchmark, accuracy on MTL15, and micro-weighted F1 scores on WikiAnn. On the MTL15 benchmark, we select 1000 random samples per class for training each task and hold out 500 samples per class for validation.\footnote{All design choices of \moclp\ are kept consistent with previous works \cite{wang-etal-2023-gradsim, wang2023orthogonal, wang-etal-2024-rehearsal} to ensure a fair comparison.} We explore three task orders for each benchmark, adopting the same multiple task orders as the prior work. Please refer to Appendix~\ref{sec:dataset-info} for more details about the benchmarks and task orders.

\subsection{Training Details}
\label{sec:training-details}
We deploy three LMs for these datasets, in line with prior work \cite{razdaibiedina2022progressive, wang-etal-2024-rehearsal}. We use encoder-based models for AfriSenti and WikiAnn NER (AfroXLM and BERT, respectively), and the encoder-decoder model T5 for MTL15. 
Prefix-tuning is used as the task-specific modules for all deployed models. All design choices are consistent with previous works to ensure a fair comparison. The reported results represent the average performance after training on all tasks consecutively and are averaged over three random seeds. The detailed experimental settings are provided in Appendix~\ref{sec:implement-details}.

\subsection{Baselines}
\label{sec:baselines}

To compare different CL methods, we include the following baselines: (1) Sequential fine-tuning continuously fine-tunes the language model on the task sequence: (a) \textbf{Seq FT (F)} refers to all model parameters are updated (fully fine-tuning), while (b) \textbf{Seq FT} only fine-tunes the PEFT parameters; (2) \textbf{Per-task FT} trains a separate PEFT module for each task; and the parameter isolation-based methods (3) \textbf{ProgPrompt} \cite{razdaibiedina2022progressive} assigns task-specific parameters and progressively concatenates modules of all tasks to encourage knowledge transfer; (4) \textbf{EPI} \cite{wang-etal-2023-rehearsal} introduces a non-parametric task identification technique to select modules for task training and inference; (5) \textbf{O-LoRA} \cite{wang2023orthogonal} learns tasks in different low-rank vector spaces that are kept orthogonal to each other to minimize interference; and (6) \textbf{MoCL} \cite{wang-etal-2024-rehearsal} introduces a modular and compositional framework that progressively expands task-specific modules and composes the new module with existing ones to facilitate knowledge transfer. A detailed description of these methods can be found in Appendix~\ref{sec:baseline-details}.

\section{Results and Analysis}
\label{sec:exp-res}
In this section, we present and analyze our experimental results.

\subsection{Overall Results}
\label{sec:results-overall}
Table \ref{tab:overall-1} shows the performance of \moclp\ and other baseline methods on the AfriSenti and WikiAnn benchmarks. \moclp\ consistently outperforms the baselines while significantly reducing the number of trainable parameters. Using only 50\% and 30\% of the trainable parameters compared to other CL methods on Afrisenti and Wikiann respectively, \moclp\ showcases an exceptional balance of efficiency and performance. 
In the MTL15 benchmark, as illustrated in Table~\ref{tab:overall-2}, \moclp\ also shows superior performance. As mentioned in prior work \cite{wang-etal-2024-rehearsal}, tasks in this benchmark share lower similarity compared to AfriSenti and WikiAnn, resulting in weaker reusability of task modules. Therefore, we do not observe a significant drop in the number of trainable parameters here as seen in the other benchmarks. However, we still achieve a 25\% reduction in parameter size while maintaining final performance.

Overall, \moclp\ demonstrates its superiority in efficiently managing the continual learning process without the substantial parameter overhead. The competitive performance of \moclp\ across different benchmarks highlights its robust adaptability and scalability to the continual learning sequence up to 176 tasks long.

\begin{table}[htbp]
  \centering
  \scalebox{0.75}{
    \begin{tabular}{l|ccccc}
    \toprule
    \multicolumn{1}{c|}{\multirow{2}[2]{*}{\textbf{Method}}} & \multicolumn{5}{c}{\textbf{MTL15}} \\
          & \textbf{AVG} & O1    & O2    & O3    & \multicolumn{1}{l}{\textbf{\# Params}} \\
    \midrule
    Seq FT-F & 7.4   & 7.4   & 7.4   & 7.5   & 770M \\
    Seq FT-P & 64.7  & 69.9  & 58.9  & 65.1  & 1.4M \\
    Per-task FT & 80.5  & 80.5  & 80.5  & 80.5  & 21.1M \\
    ProgPrompt & 77.9  & 78.0  & 77.7  & 77.9  & 21.1M \\
    EPI   & 65.4  & 62.8  & 65.3  & 68.1  & 21.1M \\
    O-LoRA & 69.6  & 78.0  & 77.7  & 77.9  & 33.8M \\
    MoCL  & \textbf{82.5}  & 82.9  & 82.8  & 81.9  & 21.1M \\
    \textbf{\moclp\ (Ours)} & \textbf{82.5}  & 83.0  & 82.7  & 81.8  & $\text{15.6M} \scriptstyle \pm\text{1.1}$ \\
    \bottomrule
    \end{tabular}%
    }
  \caption{Summary of the continual learning results on the MTL15 benchmark with the T5-large model. \moclp\ achieves the best average performance (AVG) while using fewer parameters (\# Params), demonstrating its effectiveness on the multi-task CL benchmark.}
  \label{tab:overall-2}%
\end{table}%

\begin{table}[!h]
  \centering
  \footnotesize
  \scalebox{0.92}{
    \begin{tabular}{lcccc}
    \toprule
          & \multicolumn{2}{c}{\textbf{AfriSenti}} & \multicolumn{2}{c}{\textbf{WikiAnn}} \\
          & \textbf{AVG} & \textbf{\# Params} & \textbf{AVG} & \textbf{\# Params} \\
    \midrule
    Per-task FT & 49.10 & 4.5M  & 71.22 & 24.8M \\
    \midrule
    MoCL  & 59.31 & 4.5M  & 73.80 & 24.9M \\
    \textit{w/ Gaussian} & 42.25 & 4.5M  & 67.38 & 24.8M \\
    \textit{w/ Embed mean} & 52.63 & 4.5M  & 70.12 & 24.8M \\
    \midrule
    \textbf{\moclp\ (Ours)} & \textbf{59.41} & $\text{2.2M}\scriptstyle \pm\text{0.4}$ & \textbf{73.91} & $\text{8.0M}\scriptstyle \pm\text{0.1}$ \\
    \textit{w/ Gaussian} & 42.15 & $\text{4.1M}\scriptstyle \pm\text{0.0}$ & 67.46 & $\text{5.4M}\scriptstyle \pm\text{0.3}$ \\
    \textit{w/ Embed mean} & 52.13 & $\text{3.9M}\scriptstyle \pm\text{0.2}$ & 70.21 & $\text{20.3M}\scriptstyle \pm\text{0.7}$ \\
    \bottomrule
    \end{tabular}%
    }
  \caption{Performance comparison of different task representation methods for module composition and pruning. Specifically, we use Gaussian distribution to model the input embeddings of each task (\textit{w/ Gaussian rep}) and calculate the mean of the input embeddings of each task (\textit{w/ Embed mean}). Notably, the use of these variations results in substantially lower performance compared to the original MoCL and \moclp\ which utilizes learnable feature vectors as task representations.}
  \label{tab:res-task-rep-ablation}%
\end{table}%

\begin{figure*}[!h]
    \centering
    \includegraphics[width=1\linewidth]{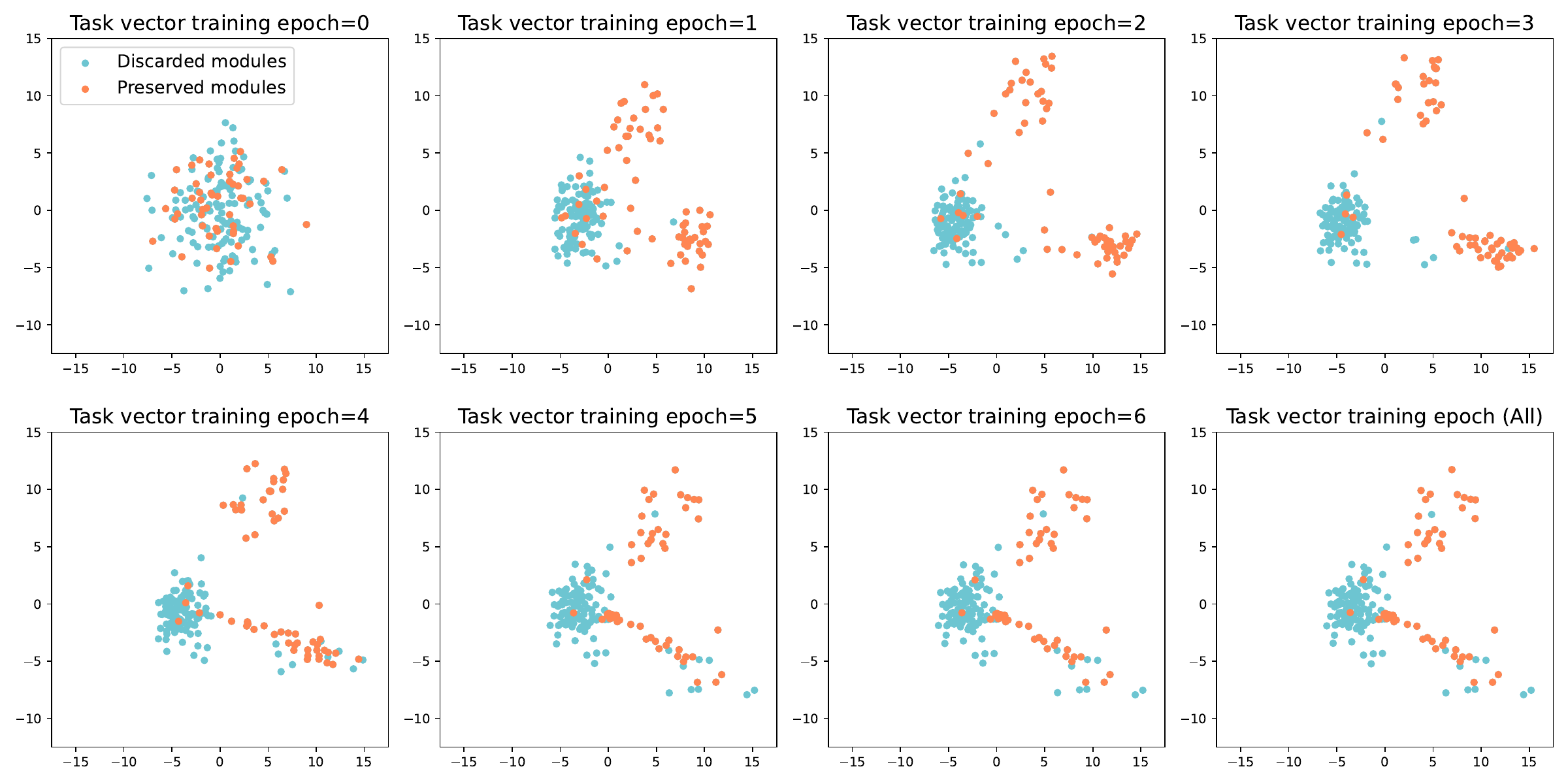}
    \caption{Visualization of task feature vectors on the WikiAnn benchmark using PCA for dimensionality reduction. We vary the training epochs for task feature vectors in \moclp. As the training epochs increase, the task feature vectors spread out from the initial dense cluster. Notably, the feature vectors of preserved modules (in orange) are spread across a large area while the discarded modules (in blue) form a dense cluster, indicating their redundancy.}
    \label{fig:task_vector_evolution}
\end{figure*}

\subsection{Task Representation Comparison}
\label{sec:res-rep-comparison}
In this work, we adopt learnable task feature vectors as task representation, and based on these, we perform module composition and pruning. In Section~\ref{sec:results-overall}, we demonstrate the effectiveness of this design choice. While this is not the only option for task representations, in this section, we experiment with two other types of task representation: (1) using Gaussian distributions to model the input embeddings of each task (\textit{w/ Gaussian}) and (2) calculating the mean of the input embeddings of each task for task representations(\textit{w/ Embed mean}).

Table~\ref{tab:res-task-rep-ablation} provides the results of using different task representation options for module composition and pruning. A significant performance drop occurs when using Gaussian distributions or the mean of task input embeddings as the task representation. In most cases, their performance is worse than the Per-Task FT baseline, indicating that using these task representations for module composition leads to performance degradation rather than beneficial knowledge transfer across tasks.

We believe that this degradation is due to the fact that both of these task representations are static and are solely based on the input embeddings. In contrast, \moclp\ utilizes trainable task feature vectors, meaning the model can automatically learn to capture the salient task features necessary for effective module composition. Trainable task representations are a better choice because not all information in the input embedding is relevant for module composition. To effectively capture reusability between task modules, the model must focus on the salient features while ignoring irrelevant ones. Static task representations, which are purely based on input embeddings, fail to achieve this selective focus.

\subsection{Ablation Study: Varying the Training Epochs for Task Feature Vectors}

\begin{figure}[!h]
    \centering
    \includegraphics[width=\linewidth]{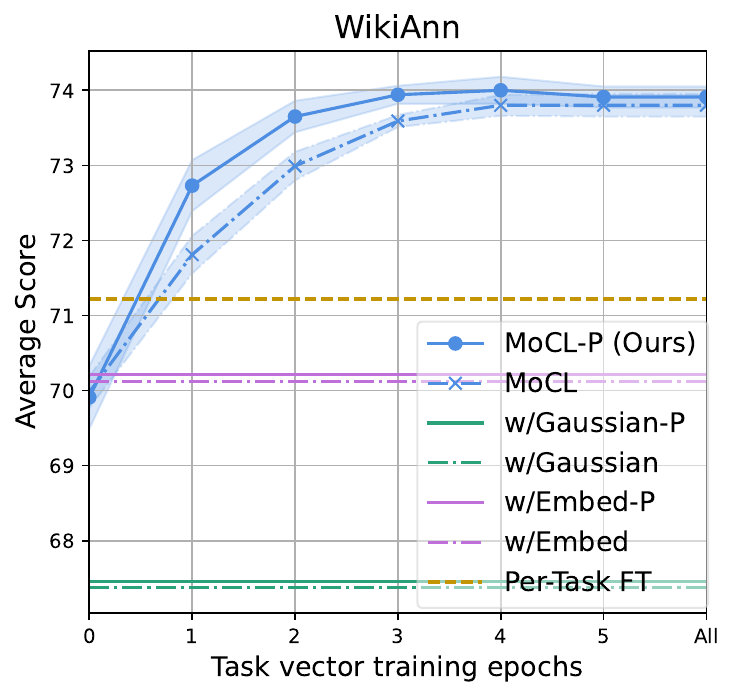}
    \caption{Experimental results with different training epochs for task feature vectors in \moclp. The model performance improves rapidly with the initial increase of the number of training epochs and converges towards a performance plateau after epochs $>4$. \moclp\ achieves significantly better performance than other task representation options. It highlights the advantage of trainable task representations in capturing salient task features for effective module composition.}
    \label{fig:res-task-vec-epoch-ablation}
\end{figure}

To substantiate our assumption introduced in Section~\ref{sec:res-rep-comparison}, we additionally conduct ablation experiments on WikiAnn where we vary the training epochs for task feature vectors in \moclp. As illustrated in Figure~\ref{fig:res-task-vec-epoch-ablation}, training the task feature vectors for different epochs shows a clear pattern: the model performance improves significantly with the initial increase of the number of training epochs. Beyond a certain point (epoch = 4), additional training does not yield further benefits and converges towards a performance plateau. This observation suggests that by allowing the model to adapt these vectors over several epochs, \moclp\ can more accurately identify and leverage the most relevant features for module composition. This underscores the critical importance of the trainable nature of task feature vectors in \moclp.

In Figure~\ref{fig:task_vector_evolution}, we visualize the task feature vectors at different training epochs on the WikiAnn dataset, which includes a total of 176 tasks. The colors represent two categories of task modules: those that are eventually discarded (blue) and those that are preserved (orange) through the learning process. Initially (training epoch = 0), the vectors are evenly distributed around the origin since they are uniformly initialized. As the training epochs increase, the task vectors spread out and become more distinct, suggesting that the model captures distinct features of tasks and utilizes them for module pruning. Notably, the feature vectors of preserved modules are spread across a large area, while the discarded modules form a dense cluster, indicating their redundancy.
The embeddings of task feature vectors stabilize by epoch 5, indicating a convergence in the task representation learning process. These observed patterns demonstrate the effectiveness of our strategy of using trainable task representations for module composition and pruning, which helps in preserving only the most salient modules for continual learning.

\begin{figure*}[!t]
    \centering
    \includegraphics[width=\linewidth]{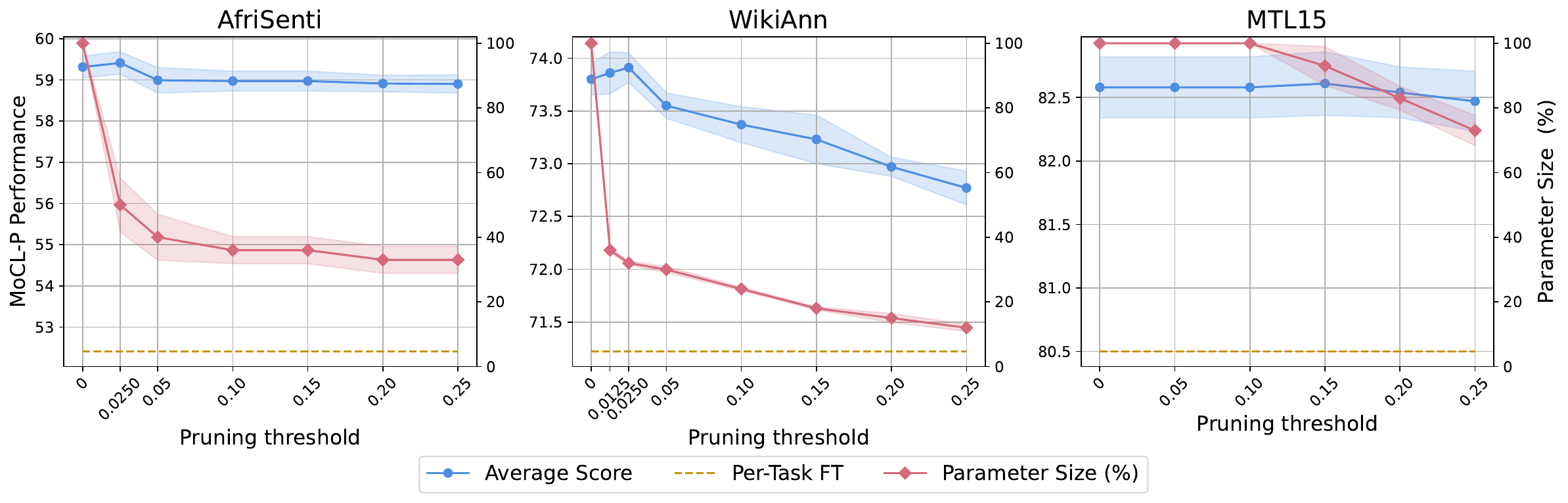}
    \caption{Impact of pruning thresholds on the performance and parameter size of \moclp\ across different benchmarks. The performance of \moclp\ exhibits robustness to different thresholds, with a significant reduction in the number of trainable parameters. This demonstrates that \moclp\ can maintain effective performance despite using fewer parameters.}
    \label{fig:res-prun-ths-ablation}
\end{figure*}
\subsection{Ablation Study: Varying the Pruning Threshold} 

In this section, we study the impact of using different thresholds on the performance of \moclp. As introduced in Section~\ref{sec:method-composition-pruning}, we compare the matching weight of the newly initialized task module $\alpha_{m}$ with the pre-specified threshold $\alpha_{ths}$, if $\alpha_{m} <  \alpha_{ths} $, then we discard the newly learned module.  We vary $\alpha_{ths}$ from 0 to 0.25 for the three benchmarks used in this work. 

The results are shown in Figure~\ref{fig:res-prun-ths-ablation}. The figure illustrates how varying the pruning threshold affects both the average performance and the parameter size across different benchmarks. 

For the model performance, we observe that the initial increase in the pruning threshold leads to a performance increase on all three benchmarks. This indicates that excluding the redundant modules benefits performance. As the threshold continues to increase, the average performance on AfriSenti and MTL15 remains relatively stable, while the performance on WikiAnn drops, possibly due to the loss of information in potentially useful modules. Additionally, it is worth noting that the performance of \moclp\ is consistently and significantly better than the Per-Task FT baseline, suggesting that \moclp\ achieves effective knowledge transfer at different pruning thresholds.
 
Furthermore, for the parameter size, a significant reduction is observed as the threshold increases on all three benchmarks. This demonstrates the superiority of \moclp\ on parameter efficiency. We observe that the parameter size decreases more pronounced and faster on AfriSenti and WikiAnn, while it decreases less and more slowly on MTL15. We believe it is due to the characteristics of the benchmarks. As mentioned in Section~\ref{sec:results-overall}, tasks in this benchmark share a lower similarity, therefore, most task modules are highly specialized to these distinct tasks and cannot be discarded.
 
We choose different pruning thresholds for different benchmarks reported in Table~\ref{tab:overall-1}. For each benchmark, we select the pruning threshold that best balances performance and parameter size to report the results in Table~\ref{tab:overall-1}. Specifically, we use $\alpha_{ths}=0.025$ for AfriSenti and WikiAnn, and $\alpha_{ths}=0.25$ for MTL15. With these thresholds,  \moclp\ achieves equally good performance with only 50\%, 30\%, and 75\% of the number of trainable parameters compared to MoCL without pruning on these three benchmarks, respectively.

\section{Conclusion}

In this paper, we introduce \moclp, a novel continual learning approach that addresses the core challenges of catastrophic forgetting, knowledge transfer, and parameter efficiency in continual learning. We utilize learnable task representations for module composition and adaptive pruning, maintaining a lightweight model while achieving state-of-the-art performance across various benchmarks. Notably, \moclp\ scales effectively to long continual learning sequences, handling up to 176 tasks without compromising performance. These experimental results showcase \moclp's potential to enhance practical machine learning applications by effectively managing computational costs, thus providing a scalable and efficient solution for real-world scenarios where minimum resource requirements are crucial.

\section*{Limitations}
While \moclp\ demonstrates significant advancements in continual learning, our study has some limitations that should be addressed in future work. First, we only use the long sequence multilingual benchmark, i.e., WikiAnn with 176 tasks, in this work due to the lack of existing long sequence multi-task benchmarks. The absence of these benchmarks limits the evaluation of \moclp’s performance across diverse multi-task scenarios. Building a long sequence multi-task benchmark for continual learning would be an interesting research direction, although it is beyond the scope of this work. Second, as we follow the evaluation setup from prior works, we do not include generative tasks for evaluation. Therefore, we may not capture the potential of \moclp\ in a wider range of continual learning challenges. Including generative tasks in future evaluations would provide a more comprehensive understanding of \moclp’s capabilities.


\bibliography{anthology,custom}
\nocite{Ando2005,andrew2007scalable,rasooli-tetrault-2015}

\newpage

\appendix
\section{Appendix}
\label{sec:appendix}

\subsection{Dataset Information}
\label{sec:dataset-info}

Here we provide detailed information on the datasets used in this work. The MTL15 benchmark consists of 15 classification tasks, combining five datasets from the standard CL benchmark MTL5 (AG News, Amazon reviews, Yelp reviews, DBpedia, and Yahoo Answers) \cite{zhang2015character}, four tasks from the GLUE benchmark (MNLI, QQP, RTE, SST2) \cite{wang2018glue}, five tasks from the SuperGLUE benchmark (WiC, CB, COPA, MultiRC, BoolQ), and the IMDB movie reviews dataset \cite{maas2011learning}. Details of the MTL15 benchmark are provided in Table~\ref{tab:mtl15-dataset}. Following \newcite{wang-etal-2024-rehearsal}, we use AfriSenti \cite{muhammad2023afrisenti, wang-etal-2023-nlnde}, a multilingual sentiment analysis dataset covering 12 low-resource African languages, including Amharic (am), Algerian Arabic (dz), Hausa (ha), Igbo (ig), Kinyarwanda (kr), Moroccan Arabic (ma), Nigerian Pidgin (pcm), Mozambican Portuguese (pt), Swahili (sw), Xitsonga (ts), Twi (twi), and Yoruba (yo). Additionally, to further evaluate the module pruning capability of \moclp, we include WikiAnn, a multilingual named entity recognition (NER) dataset that covers 176 languages. The long task sequence in WikiAnn provides an adequate testbed for evaluating the pruning functionality of \moclp. Due to space constraints, we do not list the names of the 176 languages and their corresponding abbreviations. The specific language information is available at \href{https://huggingface.co/datasets/wikiann}{https://huggingface.co/datasets/wikiann}.

\begin{table*}[p]
\centering
\begin{tabular}{l|lll}
\hline
\textbf{Dataset name} & \textbf{Category} & \textbf{Task} & \textbf{Domain} \\ 
\hline
Yelp & MTL5 & sentiment analysis & Yelp reviews \\
Amazon & MTL5 & sentiment analysis & Amazon reviews \\
DBpedia & MTL5 & topic classification & Wikipedia \\
Yahoo & MTL5 & QA & Yahoo Q\&A \\
AG News & MTL5 & topic classification & news \\
MNLI & GLUE & NLI & various \\
QQP & GLUE & paraphrase detection & Quora \\
RTE & GLUE & NLI & news, Wikipedia \\
SST2 & GLUE & sentiment analysis & movie reviews \\
WiC & SuperGLUE & word sense disambiguation & lexical databases \\
CB & SuperGLUE & NLI & various \\
COPA & SuperGLUE & QA & blogs, encyclopedia \\
BoolQ & SuperGLUE & boolean QA & Wikipedia \\
MultiRC & SuperGLUE & QA & various \\
IMDB & Other & sentiment analysis & movie reviews \\
\hline
\end{tabular}
\caption{The details of 15 datasets used in the MTL15 benchmark. NLI denotes natural language inference, and QA denotes questions and answers task.}
\label{tab:mtl15-dataset}
\end{table*}
We use different task orders for each dataset to evaluate the robustness of continual learning methods against changing task orders. For the MTL15 and AfriSenti benchmarks, we follow the task orders used in prior works, while for the WikiAnn benchmarks, we generate three random task orders for evaluation. The task orders used are summarized in Table \ref{tab:task-orders}.

\subsection{Experiment Details}
In this section, we provide the implementation details for the experiments and a detailed description of the baseline methods used in this work.

\subsubsection{Implementation Details}
\label{sec:implement-details}
We use the AdamW optimizer \cite{adamw} for all experiments. We choose the same maximum sequence length and prefix length as prior work \cite{razdaibiedina2022progressive, wang-etal-2023-rehearsal}. Table \ref{tab:hyperparams} provides detailed hyperparameter choices for \moclp\ across different datasets. The training
was performed on Nvidia A100 GPUs.\footnote{All experiments ran on a carbon-neutral GPU cluster.}

\begin{table*}[htbp]
  \centering
    \begin{tabular}{l|ccc}
    \hline
          & \multicolumn{3}{c}{\textbf{Hyperparameters}} \\
          & AfriSenti-AfroXLMR & WikiAnn-BERT & MTL15-T5 \\
    \hline
    Epochs & 40    & 5   & 40 \\
    Early stop patience & 5     & N/A   & 5 \\
    Batch size & 8     & 32  & 8 \\
    Learning rate & 2e-4 & 1e-3 & 5e-2 \\
    Max. sequence length & 256   & 128 & 512 \\
    Prefix length & 16    & 8   & 10 \\
    \hline
    \end{tabular}%
  \caption{Hyperparameters used in this work across different CL experiments.}
  \label{tab:hyperparams}%
\end{table*}%

\subsubsection{Baseline Methods}
\label{sec:baseline-details}

In Section \ref{sec:exp-res}, we evaluate \moclp\ and prior continual learning methods on different benchmark datasets. Here, we provide a more detailed description of the baseline methods used in this work.

ProgPrompt \cite{razdaibiedina2022progressive}: A parameter isolation-based continual learning method that assigns task-specific parameters to avoid catastrophic forgetting. During continual learning, ProgPrompt progressively concatenates all task-specific modules to encourage forward transfer.
 
EPI \cite{wang-etal-2023-rehearsal}: A parameter isolation-based method applicable to the class-incremental learning setting (CIL), where task identities are not given during inference. EPI introduces a non-parametric task identification module that identifies tasks during testing. Given reliable task identification, the CIL performance of EPI could be comparable to TIL, where the ground truth task identities are given during inference.

O-LoRA \cite{wang2023orthogonal}: A parameter isolation-based method that learns tasks in different low-rank vector spaces that are kept orthogonal to each other to minimize interference. It mitigates catastrophic forgetting by constraining the gradient update of the current task to be orthogonal to the gradient space of past tasks. However, the orthogonality of the gradient subspace for individual tasks also limits knowledge transfer between tasks.

MoCL \cite{wang-etal-2024-rehearsal}: Introduces a modular and compositional continual learning framework to compose the new module with existing ones based on task module matching. This compositional strategy enables effective knowledge transfer by considering task interaction.

As discussed in Section~\ref{sec:related-work}, we build on the idea of MoCL \cite{wang-etal-2024-rehearsal} by utilizing task representations for module composition, ensuring that the model effectively reuses relevant knowledge from previous tasks. Beyond that, we introduce an adaptive pruning strategy to keep the language model lightweight and effective throughout the continual learning process, making it scalable for continual learning scenarios with long task sequences.

\onecolumn
\begin{longtable}{ccc>{\centering\arraybackslash}p{10cm}} 
\caption{The different orders of task sequences used for continual learning experiments.} \label{tab:task-orders} \\
\toprule
\textbf{Dataset} & \textbf{Order} & \textbf{Model} & \textbf{Task Sequence} \\
\midrule
\endfirsthead
\multicolumn{4}{c}%
{{\bfseries \tablename\ \thetable{} -- continued from previous page}} \\
\toprule
\textbf{Dataset} & \textbf{Order} & \textbf{Model} & \textbf{Task Sequence} \\
\midrule
\endhead
\midrule \multicolumn{4}{r}{{Continued on next page}} \\
\endfoot
\bottomrule
\endlastfoot
\multirow{3}{*}{AfriSenti} & 1 & AfroXLMR & \makecell{am $\rightarrow$ dz $\rightarrow$ ha $\rightarrow$ ig $\rightarrow$ kr $\rightarrow$ ma \\ $\rightarrow$ pcm $\rightarrow$ pt $\rightarrow$ sw $\rightarrow$ ts $\rightarrow$ twi $\rightarrow$ yo} \\
& 2 & AfroXLMR & \makecell{ma $\rightarrow$ pcm $\rightarrow$ kr $\rightarrow$ pt $\rightarrow$ ig $\rightarrow$ sw \\ $\rightarrow$ ha $\rightarrow$ ts $\rightarrow$ dz $\rightarrow$ twi $\rightarrow$ am $\rightarrow$ yo} \\
& 3 & AfroXLMR & \makecell{am $\rightarrow$ dz $\rightarrow$ ha $\rightarrow$ ma $\rightarrow$ ig $\rightarrow$ kr \\ $\rightarrow$ sw $\rightarrow$ ts $\rightarrow$ twi $\rightarrow$ yo $\rightarrow$ pcm $\rightarrow$ pt} \\
\midrule
\multirow{3}{*}{MTL15} & 1 & T5 & \makecell{mnli $\rightarrow$ cb $\rightarrow$ wic $\rightarrow$ copa $\rightarrow$ qqp $\rightarrow$ boolq $\rightarrow$ rte $\rightarrow$ imdb $\rightarrow$ \\ yelp $\rightarrow$ amazon $\rightarrow$ sst2 $\rightarrow$ dbpedia $\rightarrow$ ag $\rightarrow$ multirc $\rightarrow$ yahoo} \\
& 2 & T5 & \makecell{multirc $\rightarrow$ boolq $\rightarrow$ wic $\rightarrow$ mnli $\rightarrow$ cb $\rightarrow$ copa $\rightarrow$ qqp $\rightarrow$ rte $\rightarrow$ \\ imdb $\rightarrow$ sst2 $\rightarrow$ dbpedia $\rightarrow$ ag $\rightarrow$ yelp $\rightarrow$ amazon $\rightarrow$ yahoo} \\
& 3 & T5 & \makecell{yelp $\rightarrow$ amazon $\rightarrow$ mnli $\rightarrow$ cb $\rightarrow$ copa $\rightarrow$ qqp $\rightarrow$ rte $\rightarrow$ imdb $\rightarrow$ \\ sst2 $\rightarrow$ dbpedia $\rightarrow$ ag $\rightarrow$ yahoo $\rightarrow$ multirc $\rightarrow$ boolq $\rightarrow$ wic} \\
\midrule
\multirow{3}{*}{WikiAnn} & 1 & BERT & \makecell{ga $\rightarrow$ fi $\rightarrow$ sco $\rightarrow$ bs $\rightarrow$ co $\rightarrow$ pnb $\rightarrow$ eu $\rightarrow$ vls $\rightarrow$ os $\rightarrow$ de $\rightarrow$ \\ 
hy $\rightarrow$ mwl $\rightarrow$ ca $\rightarrow$ or $\rightarrow$ wa $\rightarrow$ rw $\rightarrow$ simple $\rightarrow$ tl $\rightarrow$ crh $\rightarrow$ \\
lij $\rightarrow$ min $\rightarrow$ ko $\rightarrow$ scn $\rightarrow$ an $\rightarrow$ mk $\rightarrow$ hi $\rightarrow$ ug $\rightarrow$ ext $\rightarrow$ sl $\rightarrow$ \\
sw $\rightarrow$ nap $\rightarrow$ et $\rightarrow$ wuu $\rightarrow$ uz $\rightarrow$ mzn $\rightarrow$ ast $\rightarrow$ jv $\rightarrow$ su $\rightarrow$ \\
ilo $\rightarrow$ csb $\rightarrow$ cdo $\rightarrow$ tk $\rightarrow$ ckb $\rightarrow$ lv $\rightarrow$ ur $\rightarrow$ th $\rightarrow$ am $\rightarrow$ kn $\rightarrow$ \\
pms $\rightarrow$ ba $\rightarrow$ tt $\rightarrow$ pl $\rightarrow$ vec $\rightarrow$ ru $\rightarrow$ cs $\rightarrow$ ne $\rightarrow$ bn $\rightarrow$ es $\rightarrow$ \\
fy $\rightarrow$ fiu-vro $\rightarrow$ bo $\rightarrow$ mt $\rightarrow$ fr $\rightarrow$ mr $\rightarrow$ nn $\rightarrow$ bar $\rightarrow$ ang  $\rightarrow$ \\
no $\rightarrow$ fo $\rightarrow$ el $\rightarrow$ qu $\rightarrow$ fa $\rightarrow$ eml $\rightarrow$ kk $\rightarrow$ tr $\rightarrow$ pt $\rightarrow$ km $\rightarrow$ \\
dv $\rightarrow$ hsb $\rightarrow$ rm $\rightarrow$ ta $\rightarrow$ fur $\rightarrow$ war $\rightarrow$ frr $\rightarrow$ ps $\rightarrow$ io $\rightarrow$ da $\rightarrow$ \\
zh-yue $\rightarrow$ ms $\rightarrow$ cv $\rightarrow$ diq $\rightarrow$ mn $\rightarrow$ lb $\rightarrow$ cy $\rightarrow$ sa $\rightarrow$ ig $\rightarrow$ \\
oc $\rightarrow$ hu $\rightarrow$ arc $\rightarrow$ ln $\rightarrow$ ku $\rightarrow$ hr $\rightarrow$ nds $\rightarrow$ az $\rightarrow$ ar $\rightarrow$ ce $\rightarrow$ \\
lt $\rightarrow$ zea $\rightarrow$ it $\rightarrow$ zh-classical $\rightarrow$ be-x-old $\rightarrow$ mi $\rightarrow$ ia $\rightarrow$ is  $\rightarrow$ \\ 
la $\rightarrow$ sv $\rightarrow$ nl $\rightarrow$ gd $\rightarrow$ pa $\rightarrow$ xmf $\rightarrow$ ksh $\rightarrow$ zh-min-nan $\rightarrow$ \\
lmo $\rightarrow$ tg $\rightarrow$ sh $\rightarrow$ eo $\rightarrow$ zh $\rightarrow$ te $\rightarrow$ he $\rightarrow$ vep $\rightarrow$ as $\rightarrow$ \\
yi $\rightarrow$ cbk-zam $\rightarrow$ yo $\rightarrow$ ro $\rightarrow$ ace $\rightarrow$ id $\rightarrow$ jbo $\rightarrow$ nov $\rightarrow$ bg $\rightarrow$ \\
map-bms $\rightarrow$ be $\rightarrow$ sr $\rightarrow$ sah $\rightarrow$ ml $\rightarrow$ my $\rightarrow$ vo $\rightarrow$ so $\rightarrow$ gu $\rightarrow$ \\ 
br $\rightarrow$ gl $\rightarrow$ ka $\rightarrow$ li $\rightarrow$ pdc $\rightarrow$ ky $\rightarrow$ bat-smg $\rightarrow$ als $\rightarrow$ mg $\rightarrow$ \\
szl $\rightarrow$ gn $\rightarrow$ ceb $\rightarrow$ vi $\rightarrow$ sq $\rightarrow$ mhr $\rightarrow$ ay $\rightarrow$ en $\rightarrow$ bh $\rightarrow$ uk $\rightarrow$ \\
gan $\rightarrow$ sk $\rightarrow$ si $\rightarrow$ hak $\rightarrow$ af $\rightarrow$ ja $\rightarrow$ arz $\rightarrow$ sd} \\
& 2 & BERT & \makecell{
wuu $\rightarrow$ cy $\rightarrow$ mwl $\rightarrow$ eu $\rightarrow$ gn $\rightarrow$ scn $\rightarrow$ ka $\rightarrow$ pdc $\rightarrow$ it $\rightarrow$ \\ 
ro $\rightarrow$ pnb $\rightarrow$ ig $\rightarrow$ tl $\rightarrow$ sah $\rightarrow$ is $\rightarrow$ ga $\rightarrow$ ml $\rightarrow$ wa $\rightarrow$ \\
vo $\rightarrow$ simple $\rightarrow$ hr $\rightarrow$ dv $\rightarrow$ mn $\rightarrow$ csb $\rightarrow$ sl $\rightarrow$ gl $\rightarrow$ fy $\rightarrow$ \\
bn $\rightarrow$ tg $\rightarrow$ fr $\rightarrow$ th $\rightarrow$ vls $\rightarrow$ arz $\rightarrow$ zh-classical $\rightarrow$ ln $\rightarrow$ tr $\rightarrow$ \\
su $\rightarrow$ min $\rightarrow$ si $\rightarrow$ ur $\rightarrow$ sr $\rightarrow$ et $\rightarrow$ eo $\rightarrow$ sh $\rightarrow$ li $\rightarrow$ \\
fiu-vro $\rightarrow$ rw $\rightarrow$ no $\rightarrow$ mg $\rightarrow$ mr $\rightarrow$ oc $\rightarrow$ nap $\rightarrow$ yi $\rightarrow$ pa $\rightarrow$ \\
lt $\rightarrow$ ug $\rightarrow$ co $\rightarrow$ tt $\rightarrow$ sv $\rightarrow$ uk $\rightarrow$ so $\rightarrow$ ext $\rightarrow$ ky $\rightarrow$ ru $\rightarrow$ \\
kk $\rightarrow$ sa $\rightarrow$ la $\rightarrow$ el $\rightarrow$ hsb $\rightarrow$ be-x-old $\rightarrow$ bg $\rightarrow$ pt $\rightarrow$ bh $\rightarrow$ \\
br $\rightarrow$ mt $\rightarrow$ ne $\rightarrow$ id $\rightarrow$ te $\rightarrow$ cv $\rightarrow$ fo $\rightarrow$ cdo $\rightarrow$ bs $\rightarrow$ \\
lij $\rightarrow$ sw $\rightarrow$ he $\rightarrow$ ceb $\rightarrow$ hak $\rightarrow$ es $\rightarrow$ kn $\rightarrow$ mk $\rightarrow$ am $\rightarrow$ \\
or $\rightarrow$ ms $\rightarrow$ az $\rightarrow$ als $\rightarrow$ my $\rightarrow$ ce $\rightarrow$ os $\rightarrow$ ca $\rightarrow$ tk $\rightarrow$ diq $\rightarrow$ \\
zh $\rightarrow$ fi $\rightarrow$ jbo $\rightarrow$ mhr $\rightarrow$ ay $\rightarrow$ pms $\rightarrow$ rm $\rightarrow$ zea $\rightarrow$ en $\rightarrow$ \\
zh-yue $\rightarrow$ sco $\rightarrow$ ang $\rightarrow$ bo $\rightarrow$ ar $\rightarrow$ ia $\rightarrow$ zh-min-nan $\rightarrow$ ckb $\rightarrow$ \\
fa $\rightarrow$ crh $\rightarrow$ as $\rightarrow$ yo $\rightarrow$ szl $\rightarrow$ fur $\rightarrow$ hi $\rightarrow$ eml $\rightarrow$ mi $\rightarrow$ lb $\rightarrow$ \\
de $\rightarrow$ bat-smg $\rightarrow$ uz $\rightarrow$ lv $\rightarrow$ nov $\rightarrow$ ast $\rightarrow$ cs $\rightarrow$ hy $\rightarrow$ sk $\rightarrow$ \\
sq $\rightarrow$ be $\rightarrow$ xmf $\rightarrow$ af $\rightarrow$ ps $\rightarrow$ qu $\rightarrow$ da $\rightarrow$ ja $\rightarrow$ vep $\rightarrow$ ku $\rightarrow$ \\
mzn $\rightarrow$ nl $\rightarrow$ vec $\rightarrow$ map-bms $\rightarrow$ ace $\rightarrow$ io  $\rightarrow$ gu $\rightarrow$ bar $\rightarrow$ \\
ilo $\rightarrow$ km $\rightarrow$ arc $\rightarrow$ cbk-zam $\rightarrow$ pl $\rightarrow$ ksh $\rightarrow$ war $\rightarrow$ gd $\rightarrow$ ba $\rightarrow$ \\
lmo $\rightarrow$ gan $\rightarrow$ ko $\rightarrow$ an $\rightarrow$ frr $\rightarrow$ vi $\rightarrow$ hu $\rightarrow$ jv $\rightarrow$ sd $\rightarrow$ \\
nds $\rightarrow$ nn $\rightarrow$ tas} \\
& 3 & BERT & \makecell{
tl $\rightarrow$ sah $\rightarrow$ ckb $\rightarrow$ qu $\rightarrow$ az $\rightarrow$ ast $\rightarrow$ mr $\rightarrow$ eo $\rightarrow$ wa $\rightarrow$ \\
zh-classical $\rightarrow$ fiu-vro $\rightarrow$ eu $\rightarrow$ nl $\rightarrow$ map-bms $\rightarrow$ id $\rightarrow$ szl $\rightarrow$ \\
mi $\rightarrow$ io $\rightarrow$ lt $\rightarrow$ war $\rightarrow$ my $\rightarrow$ bat-smg $\rightarrow$ jv $\rightarrow$ en $\rightarrow$ \\
zh-min-nan $\rightarrow$ sh $\rightarrow$ su $\rightarrow$ frr $\rightarrow$ am $\rightarrow$ hu $\rightarrow$ hy $\rightarrow$ zh $\rightarrow$ ps $\rightarrow$ \\
hi $\rightarrow$ tg $\rightarrow$ pl $\rightarrow$ nov $\rightarrow$ dv $\rightarrow$ min $\rightarrow$ jbo $\rightarrow$ diq $\rightarrow$ ksh $\rightarrow$ \\
gn $\rightarrow$ vec $\rightarrow$ nds $\rightarrow$ lij $\rightarrow$ pdc $\rightarrow$ os $\rightarrow$ rw $\rightarrow$ als $\rightarrow$ sq $\rightarrow$ \\
fi $\rightarrow$ da $\rightarrow$ sr $\rightarrow$ ru $\rightarrow$ uz $\rightarrow$ fr $\rightarrow$ scn $\rightarrow$ tt $\rightarrow$ bh $\rightarrow$ bn $\rightarrow$ \\
mwl $\rightarrow$ et $\rightarrow$ hsb $\rightarrow$ kn $\rightarrow$ rm $\rightarrow$ nn $\rightarrow$ mhr $\rightarrow$ bg $\rightarrow$ sd $\rightarrow$ \\
ko $\rightarrow$ la $\rightarrow$ ka $\rightarrow$ de $\rightarrow$ he $\rightarrow$ pt $\rightarrow$ cs $\rightarrow$ hr $\rightarrow$ tk $\rightarrow$ cy $\rightarrow$ \\
co $\rightarrow$ or $\rightarrow$ csb $\rightarrow$ bar $\rightarrow$ mt $\rightarrow$ vo $\rightarrow$ oc $\rightarrow$ simple $\rightarrow$ ml $\rightarrow$ \\
bs $\rightarrow$ km $\rightarrow$ sk $\rightarrow$ ang $\rightarrow$ br $\rightarrow$ xmf $\rightarrow$ ay $\rightarrow$ zea $\rightarrow$ ln $\rightarrow$ \\
sco $\rightarrow$ ku $\rightarrow$ ilo $\rightarrow$ lv $\rightarrow$ mzn $\rightarrow$ zh-yue $\rightarrow$ gan $\rightarrow$ ta $\rightarrow$ gl $\rightarrow$ \\
ca $\rightarrow$ hak $\rightarrow$ mg $\rightarrow$ ne $\rightarrow$ ur $\rightarrow$ cbk-zam $\rightarrow$ uk $\rightarrow$ mn $\rightarrow$ fy $\rightarrow$ \\
ba $\rightarrow$ nap $\rightarrow$ kk $\rightarrow$ yo $\rightarrow$ tr $\rightarrow$ so $\rightarrow$ fo $\rightarrow$ ug $\rightarrow$ ace $\rightarrow$ fur $\rightarrow$ \\
pa $\rightarrow$ lmo $\rightarrow$ it $\rightarrow$ be-x-old $\rightarrow$ sa $\rightarrow$ arc $\rightarrow$ ig $\rightarrow$ lb $\rightarrow$ ms $\rightarrow$ \\
th $\rightarrow$ cv $\rightarrow$ arz $\rightarrow$ bo $\rightarrow$ el $\rightarrow$ eml $\rightarrow$ gd $\rightarrow$ pnb $\rightarrow$ cdo $\rightarrow$ \\
ky $\rightarrow$ af $\rightarrow$ vls $\rightarrow$ be $\rightarrow$ ga $\rightarrow$ es $\rightarrow$ yi $\rightarrow$ si $\rightarrow$ ext $\rightarrow$ gu $\rightarrow$ \\
mk $\rightarrow$ ja $\rightarrow$ is $\rightarrow$ no $\rightarrow$ ceb $\rightarrow$ ro $\rightarrow$ sv $\rightarrow$ ar $\rightarrow$ an $\rightarrow$ te $\rightarrow$ \\
sl $\rightarrow$ sw $\rightarrow$ wuu $\rightarrow$ pms $\rightarrow$ fa $\rightarrow$ vi $\rightarrow$ as $\rightarrow$ ce $\rightarrow$ vep $\rightarrow$ \\
li $\rightarrow$ ia $\rightarrow$ crh} \\
\end{longtable}

\end{document}